%% file: main.tex
\documentclass[10pt,twocolumn,letterpaper]{article}

\usepackage{cvpr}              

\usepackage{physics}

\usepackage{booktabs}           
\usepackage{array}
\usepackage[table]{xcolor}      
\usepackage{amsmath}            
\usepackage{pifont}
\usepackage{listings}
\usepackage{xcolor}
\usepackage{multirow}
\usepackage[accsupp]{axessibility}  
\usepackage{float}

\lstdefinelanguage{json}{
  basicstyle=\ttfamily\footnotesize,
  showstringspaces=false,
  breaklines=true,
  breakatwhitespace=false,
  columns=fullflexible,
  alsoletter=-,
  morestring=[b]",
  morecomment=[l]{//},
}
\usepackage{makecell}

\definecolor{cvprblue}{rgb}{0.21,0.49,0.74}
\usepackage[pagebackref,breaklinks,colorlinks,allcolors=cvprblue]{hyperref}

\def\paperID{3527} 
\def\confName{CVPR}
\def\confYear{2026}

\title{Fashion130K: An E-commerce Fashion Dataset for Outfit Generation with Unified Multi-modal Condition}

\author{
    Yu He \quad Ting Zhu \quad Yichun Liu \quad Lichen Ma \quad Xinyuan Shan \quad Jingling Fu \\
    Yu Shi \quad Junshi Huang\thanks{Corresponding author.} \quad Yan Li \\
    JD.com \\
    {\tt\small \{heyu2579, junshi.huang\}@gmail.com}
}

\begin{document}
\maketitle
\input{sec/0_abstract}    
\input{sec/1_intro}
\input{sec/2_related_work}
\input{sec/3_data}
\input{sec/4_method}
\input{sec/5_exp}
\input{sec/6_conclusion}
{
    \small
    \bibliographystyle{ieeenat_fullname}
    \bibliography{main}
}


\end{document}

%% file: sec/0_abstract.tex
\begin{abstract}
Recent research work on fashion outfit generation focuses on promoting visual consistency of garment by leveraging key information from reference image and text prompt. 
However, the potential of outfit generation remains underexplored, requiring comprehensive e-commercial dataset and elaborative utilization of multi-modal condition.
In this paper, we propose a brand-new e-commerce dataset, named Fashion130K, with various occasions, models, and garment types.
For the consistent generation of garment, we design a framework with Unified Multi-modal Condition (UMC) to align and integrate the text and visual prompts into generation model.
Specifically, we explore an embedding refiner to extract the unified embeddings of multi-modal prompts, within which a Fusion Transformer is proposed to align the multi-modal embeddings by adjusting the modality gap between text and image.
Based on unified embedding, the attention in generation model is redesigned to emphasis the correlations between prompts and noisy image, inducing that the noisy image can select the pivotal tokens of prompts for consistent outfit generation.
Our dataset and framework offer a general and nuanced exploration of multi-modal prompts for generation models.
Extensive experiments on real-world applications and benchmark demonstrate the effectiveness of UMC in visual consistency, achieving promising result than that of SoTA methods.
\end{abstract}

%% file: sec/1_intro.tex
\section{Introduction}
\label{sec:intro}

Diffusion models \cite{ho2020denoising,song2020denoising,rombach2022sd} have rapidly advanced image synthesis, enabling high-fidelity generation conditioned on visual and linguistic prompt. 
In the task of fashion outfit generation, recent diffusion-based methods \cite{wang2024stablegarment,choi2024improving,xu2025ootdiffusion} inject the embeddings of visual garment and text caption into attention module to improve subject consistency.
However, they still strive for well-designed information representation and integration.
The strategy of model ensemble can enhance the representation capability of conditions, while the alignment of multi-modal condition is still underexplored.

\begin{figure}[t]
    \centering
    \includegraphics[width=\linewidth]{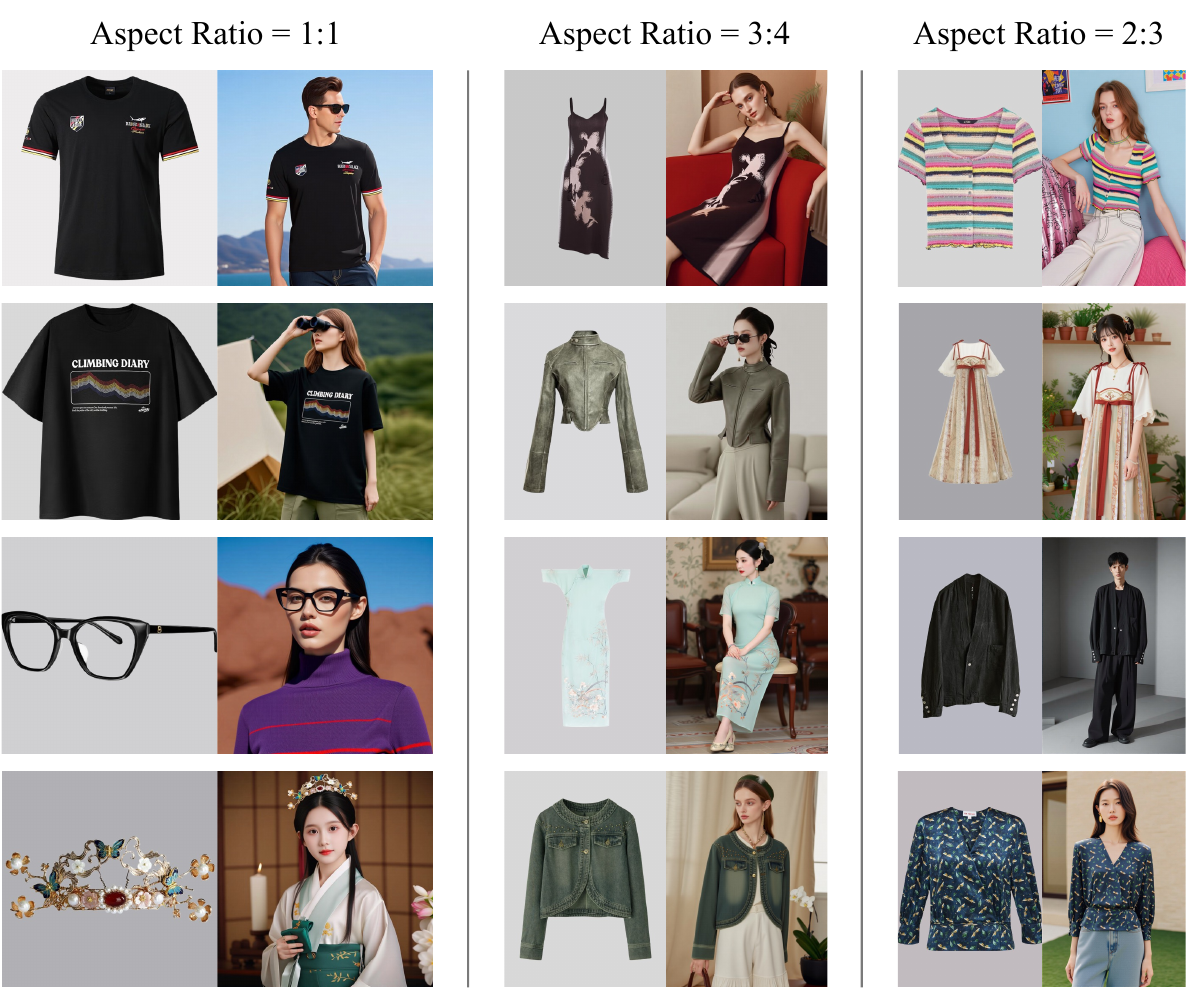}
    \caption{Our UMC model enables fashion outfit generation with various aspect ratios (1:1, 3:4, and 2:3).}
    \label{fig:demo}
\end{figure}

The quality of dataset is crucial to the performance of outfit generation. 
Widely used benchmarks such as VITON-HD \cite{choi2021viton} and DressCode \cite{morelli2022dress} advance the development of image generation but is still deficient in multiple resolutions, real-world occasions and various models.
DressCode \cite{morelli2022dress} focuses on the construction of fine-grained clothing categories with only full-body models and standing pose.
VITON-HD \cite{choi2021viton} standardizes images at fixed resolution with studio-like backgrounds, which simplifies training while reduces variety of occasion.
These limitations hinder the performance of outfit generation in real-world applications. 
Recent works \cite{han2024emma} further conclude that effective utilization of multi-modal conditions remains open and non-trivial.

In this paper, we aim to address the challenges of dataset and modeling of multi-modal condition in the task of fashion outfit generation.
We propose UMC, a diffusion-based framework that (i) aligns representation space of visual and text embeddings while maintaining the intrinsic property of modalities, and (ii) strengthens token correlations through a Selective Attention that reinforces noised tokens toward the most relevant condition tokens. 
To advance research in the fashion generation community, we release Fashion130K, a new open-source dataset of 130k model-garment image pairs with structured prompt curated from e-commerce platform. 
Unlike prior benchmarks, Fashion130K features rich background, fine-grained category coverage including accessories (such as watches, eyewear, and jewelry), and multiple aspect-ratio samples that benefits post-training.
The structured prompt describes occasion and model profile but deliberately omit garment details, ensuring that the garment appearance solely refer to the reference garment image.
On Fashion130K, UMC establishes SoTA results and shows clear qualitative advantages in outfit generation.
Overall, UMC and Fashion130K offer a practical foundation for advancing outfit generation toward realistic, diverse, and text-aware e-commerce scenario, while remaining compatible to emerging diffusion-based methods.

In summary, our work makes the following contributions:
\begin{itemize}
    \item We release Fashion130K which comprises 130k multi-resolution model-garment image pairs with diverse occasions, garment types and model poses.
    This turns Fashion130K into a standard benchmark that enables garment-text-driven and multi-resolution image generation in realistic e-commerce scenes. 
    \item We explore various structures of Embedding Refiner and propose the Fusion Transformer to align the latent space of text and image. 
    Based on the divide-and-merge structure, our Embedding Refiner can appropriately bridge the modality gap without sacrificing intrinsic property of modality, leading to more effective multi-modal embeddings. 
    \item We redesign the integration of multi-modal embeddings by employing Selective Attention in DiT blocks to highlight the pivotal condition tokens for details preservation. These innovations achieve compelling visual fidelity and consistency on Fashion130K and VITON-HD \cite{choi2021viton} without compromise on diversity and fidelity. 
\end{itemize}

%% file: sec/2_related_work.tex
\section{Related Work}
\label{sec:rel_work}

\subsection{Text-to-image Generation}
Text-to-image (T2I) generation \cite{saharia2022photorealistic, baldridge2024imagen, team2024kolors, qin2025lumina} has rapidly evolved with the advent of diffusion models, significantly surpassing previous methods based on generative adversarial networks (GANs) \cite{goodfellow2014generative,brock2018large,karras2019style} in both realism and fidelity. Latent Diffusion Models (LDM) further improved efficiency and scalability by operating directly within latent embedding spaces, thereby enabling higher-resolution image synthesis at reduced computational costs. Landmark diffusion-based approaches such as Stable Diffusion series \cite{rombach2022sd,podell2023sdxl,esser2024sd3}, DALL-E 2 \cite{ramesh2022dalle2}, and PixArt-$\alpha$ \cite{chen2023pixart} have demonstrated the ability to synthesize diverse, high-quality images from textual descriptions. Recently, DiTs have emerged as the new standard, exemplified by models such as Hunyuan-DiT \cite{li2024hunyuan} which advances the DiT framework by incorporating a multi-resolution design and enhancing fine-grained semantic alignment, especially in Chinese-centric scenarios. FLUX \cite{batifol2025flux} further allows unified and sequential processing of multi-modal conditions. Despite these advancements, purely text-driven generation still faces challenges in accurately preserving detailed visual attributes, highlighting the necessity of introducing complementary conditioning signals.

\subsection{Conditional Image Generation}
Conditional image generation integrates visual references with text or other guidance signals to precisely control specific attributes of the generated images, with particular emphasis on maintaining visual consistency such as identity, appearance, and detailed textures. Early methods like DreamBooth \cite{ruiz2023dreambooth} customize identity through subject-specific fine-tuning, but face limitations in scalability. IP-Adapter \cite{ye2023ip} addresses this by directly injecting visual features into pretrained diffusion models, reducing retraining efforts. Additionally, identity-focused methods such as PuLID \cite{guo2024pulid} introduce contrastive alignment strategies to efficiently encode and preserve identity features across generated images. In line with recent advances, unified frameworks like Omnigen2 \cite{wu2025omnigen2} and Qwen Image Edit \cite{wu2025qwen} exploit extensive multi-modal training to enhance consistency across diverse conditions. In parallel, virtual try-on tasks \cite{wang2024stablegarment, choi2024improving, li2024anyfit} have been extensively studied, with methods like Leffa \cite{zhou2025learning} introducing learnable flow fields in attention mechanisms to enhance controllability and visual realism. However, these approaches often struggle when multiple modalities interact, causing inaccuracies in fine details of the garments such as texture, logos, and structures. Our proposed UMC overcomes these limitations by explicitly fusing image and text embeddings and employing selective attention to preserve critical condition attributes, thus significantly improving consistency compared to Omnigen2, Qwen Image Edit, and SoTA outfit generation methods.

%% file: sec/3_data.tex
\section{Fashion130K Dataset}
\label{sec:data}

\begin{table*}[tbp]
\centering
\setlength{\tabcolsep}{5pt}
\renewcommand{\arraystretch}{1.05}
\begin{tabular}{@{}l cccccc@{}}
\toprule
\textbf{Dataset} & \textbf{\makecell[c]{Fine-Grained\\Category}} & \textbf{\makecell[c]{Context-Rich\\Occasion}} & \textbf{\makecell[c]{Multiple Aspect \\Ratio/Resolution}} & \textbf{\makecell[c]{Diverse Structured\\Caption}} & \textbf{\# Images} & \textbf{\# Garments} \\

\midrule
\rowcolor{gray!12}
\multicolumn{7}{@{}l@{}}{\textit{Non-Public}} \\
O-VITON \cite{neuberger2020image}        & \ding{51} & \ding{55} & \ding{55} & \ding{55} & $52,000$ & - \\
TryOnGAN \cite{lewis2021tryongan}        & \ding{51} & \ding{55} & \ding{55} & \ding{55} & $105,000$ & - \\
Revery AI \cite{li2021toward}            & \ding{51} & \ding{55} & \ding{55} & \ding{55} & $642,000$ & $321,000$ \\
Zalando \cite{yildirim2019generating}    & \ding{51} & \ding{55} & \ding{55} & \ding{55} & $1,520,000$ & $1,140,000$ \\

\midrule
\rowcolor{gray!12}
\multicolumn{7}{@{}l@{}}{\textit{Public}} \\
VITON-HD \cite{choi2021viton}            & \ding{55} & \ding{55} & \ding{55} & \ding{55} & $27,358$ & $13,679$ \\
FashionOn \cite{hsieh2019fashionon}      & \ding{55} & \ding{55} & \ding{55} & \ding{55} & $32,685$ & $10,895$ \\
DeepFashion \cite{liu2016deepfashion}    & \ding{55} & \ding{51} & \ding{55} & \ding{55} & $33,849$ & $11,283$ \\
MVP \cite{dong2019towards}               & \ding{55} & \ding{55} & \ding{55} & \ding{55} & $49,211$ & $13,524$ \\
FashionTryOn \cite{zheng2019virtually}   & \ding{55} & \ding{55} & \ding{55} & \ding{55} & $86,142$ & $28,714$ \\
LookBook \cite{yoo2016pixel}             & \ding{51} & \ding{51} & \ding{55} & \ding{55} & $84,748$ & $9,732$ \\
MVG \cite{wang2025mv}                    & \ding{55} & \ding{55} & \ding{55} & \ding{55} & $1,009$ & $5,045$ \\
VITON \cite{han2018viton}                & \ding{55} & \ding{55} & \ding{55} & \ding{55} & $32,506$ & $16,253$ \\
Dress Code \cite{morelli2022dress}       & \ding{51} & \ding{55} & \ding{55} & \ding{55} & $107,584$ & $53,792$ \\ 

\textbf{Fashion130K}                    & \ding{51} & \ding{51} & \ding{51} & \ding{51} & $\mathbf{130,386}$ & $\mathbf{130,386}$ \\ 

\bottomrule
\end{tabular}
\caption{
Overview of Fashion130K in comparison with datasets commonly used for fashion outfit generation research.
}
\label{tab:data_comparison}
\end{table*}

\begin{figure}[t]
    \centering
    \includegraphics[width=1.0\linewidth]{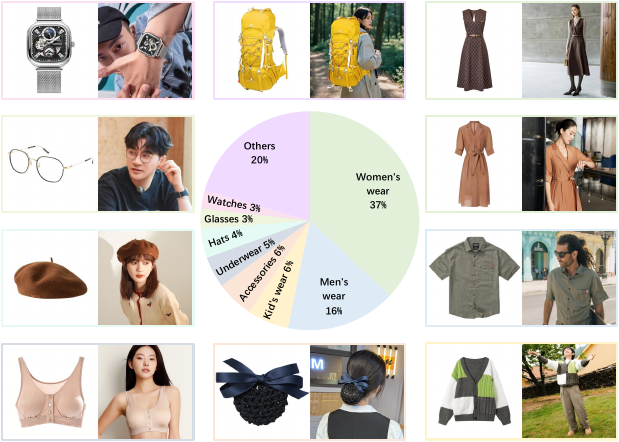}
    \caption{Category distribution and visual examples from the Fashion130K dataset.}
    \label{fig:data_dist}
\end{figure}



In this section, we introduce Fahsion130K, a large-scale fashion dataset comprising 130k e-commerce outfit images with the corresponding reference image of the garment and structured text prompt.
This dataset comprises diverse outfit images in terms of occasions, models, and garment types.
To support the post-training of generation models, we collect high quality images with aesthetic scores, high-resolution, and multiple aspect ratios.
In our study, we explore the effectiveness of plain text and structured text as prompt, and demonstrate the superior performance of structured text prompt, which is also provided in our dataset.

\subsection{Dataset Curation}

Fashion130K is built on the product image gallery collected from the e-commerce platform.
Each entry in Fashion130K includes one garment image, one model image of dressing the garment, and a structured text prompt.
To collect pairs of model-garment image, we first sample 10M candidate SKU (Stock Keeping Unit) records according to the clothing category distribution.
We de-duplicate the similar images via pretrained image feature, and remove the low-quality images with massive text or stickers by in-house OCR, sticker detectors, and aesthetic scoring model.
After that, we obtain 1.4M high potential SKU records.

To collect the matched model-garment pairs, we classify all images of SKU into garment with/without person, and obtain the most matched model-garment pair via customized image matching model.
Then, the appearance consistency of the garment in image pairs is manually annotated and 340k image pairs are obtained.
Finally, we sample 130k high-quality image pairs to ensure a balanced distribution of clothing category and background occasion.

\subsection{Diversity and Quality Analysis}
As shown in \cref{tab:data_comparison}, to our knowledge, Fashion130K is the largest open-source fashion outfit dataset, comprising $130,386$ samples of model-garment image pairs with structured caption.
The visual diversity in Fashion130K surpasses that of prior datasets.
In particular, the model images are high-quality and real-world photographs depicting models on various indoor and outdoor occasions including living-room, bedroom, street, lawn and beach \textit{et.al}.
This dataset also provides hierarchical clothing categories.
As illustrated in \cref{fig:data_dist}, Fashion130K contains 9 first-level categories and 203 second-level categories of clothing, including women’s wear, men’s wear, kid’s wear, accessories and underwear.
In addition to common categories like shirt, pant, and dress, our dataset contains fashion accessory such as watch, eyeglasses, handbag and jewelry, which are usually absent from previous fashion datasets.
The coverage of our dataset enables the comprehensive generation of fashion outfit images.

In our dataset, each pair of model-garment images has been verified by customized model and manual annotators so that the garment in the model image exactly matches the garment of the product image without severe occlusion.
We also provide multiple settings of resolutions and aspect-ratios of outfit images, such as $1024 \times 768$, $1536 \times 1024$, $1024 \times 1024$ \textit{et.al}, to prevent the degradation of multi-resolution generation in post-training.
Different from previous datasets, we carefully design a structured garment-agnostic prompt in Fashion130K.
Each prompt is composed of multiple descriptions including the background occasion, model profile, and object interaction.
Rather than using the caption of garment, we propose to use the product image of garment as visual prompt, eliminating the impact of imprecise text description.
This proposed multi-modal condition, which requires the consistent appearance of garment in model image, greatly reduces the text ambiguity and yields cleaner information for outfit generation.


%% file: sec/4_method.tex
\section{Methodology}
\label{sec:mehod}

\begin{figure*}[t]
    \centering
    \includegraphics[width=\textwidth]{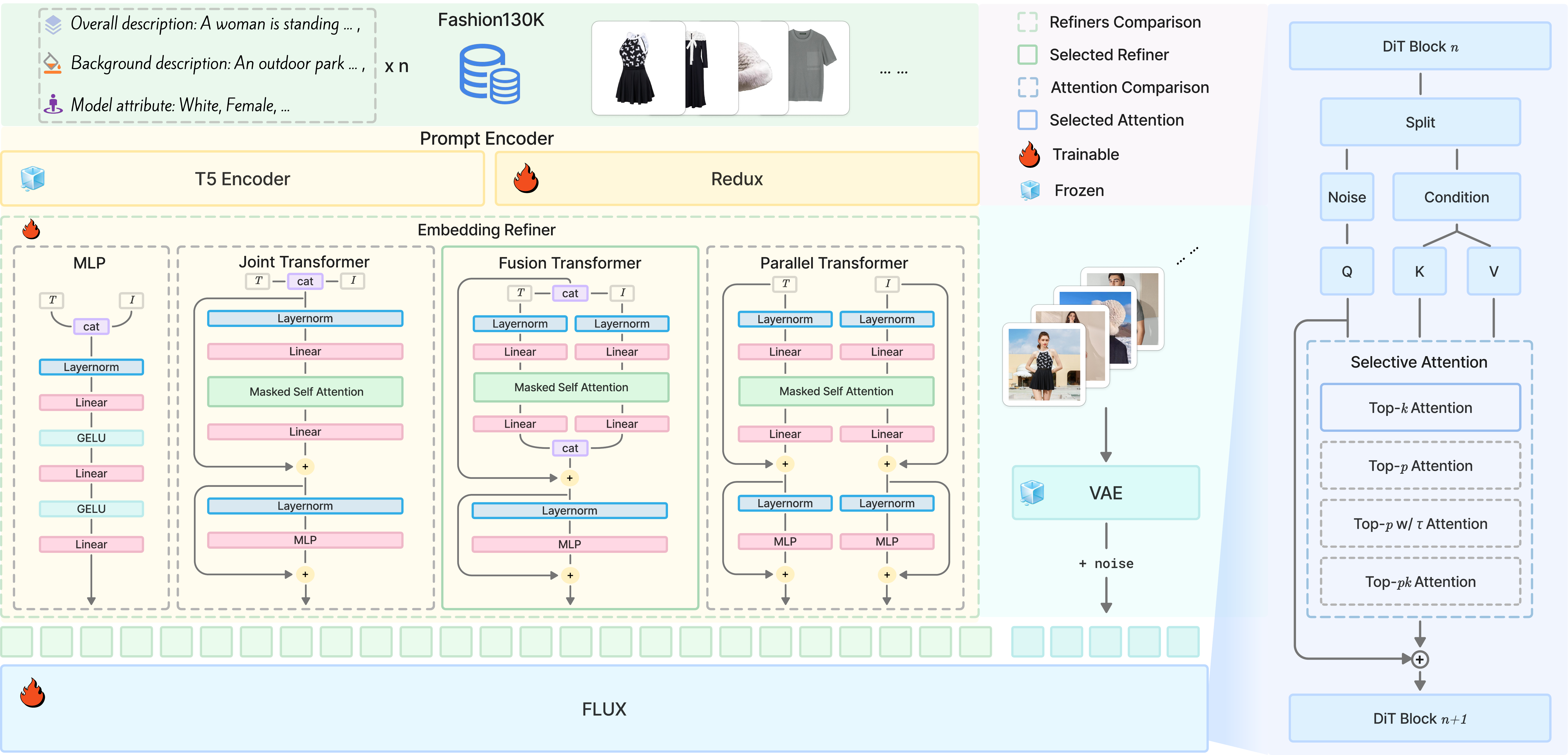}
    \caption{Overview of the UMC framework, which integrates a multi-modal Embedding Refiner and Selective Attention. We explore various structures for Embedding Refiner and select Fusion Transformer to unify the multi-modal embeddings. Meanwhile, a top-$k$ attention is used to enhance the correlation between important condition tokens and noisy image.}
    \label{fig:prc_framework}
\end{figure*}

In this section, we present the overall architecture of our proposed UMC framework as shown in ~\cref{fig:prc_framework}. 
We begin by briefly reviewing the DiT and Flow Matching (FM) methods.
In ~\cref{subsec:refiner}, the Embedding Refiner is introduced to learn a unified embedding for multi-modal conditions.
Then, we introduce the Selective Attention in ~\cref{subsec:enhancer}, which explores the strategies for elaborative matching of most relevant condition tokens and noisy image. 
Equipping with our proposed modules, the UMC framework achieves high-fidelity and more consistent images in fashion outfit generation.

\subsection{Preliminaries}
\label{subsec:pre}

DiT \cite{peebles2023dit} has emerged as a powerful generation model, combining the sequential modeling capability of Transformers with diffusion-based framework. 
Based on DiT, SD3 \cite{esser2024sd3} and MM-DiTs \cite{li2025dual} transfer multiple inputs (\textit{e.g.}, text and noise) into token sequences and progressively denoise initial Gaussian samples through iterative denoising steps. 
Recently, FM \cite{lipman2022flow, liu2022flow,holderrieth2025introduction} provides an alternative framework, directly modeling the continuous flow between a Gaussian prior and the data distribution. 
Instead of discretizing the reverse diffusion process, FM defines an explicit trajectory $z_t$ from noise to data. The latent representation at time step $t$ is defined as: $z_t=tz_0+(1-t)\epsilon$, where $z_0$ is the clean data sample and $\epsilon\sim \mathcal{N}(0,I)$ is Gaussian noise. 
This linear interpolation formulates a continuous path where $z_t$ transforms smoothly from noise ($t=0$) to data ($t=1$).
The model learns a target velocity field that is the derivative of $z_t$ to guide this flow:
\begin{equation}
    \frac{dz_t}{dt} = z_0-\epsilon
\end{equation}
Training loss minimizes the discrepancy between the learned velocity $v_\theta(z_t, t)$ and the true derivative of $z_t$:
\begin{equation}
    \mathcal{L}_{FM}(\theta) = \mathbb{E}_{t,z_0,\epsilon} \left[ \| v_\theta(z_t, t) - (z_0-\epsilon) \|_2^2 \right]
\end{equation}
The unified framework of DiT with FM provides a flexible and efficient method for image generation, improving generation fidelity and textual consistency.

\begin{figure}[t]
    \centering
    \begin{subfigure}{0.49\linewidth}
        \centering
        \includegraphics[width=\linewidth]{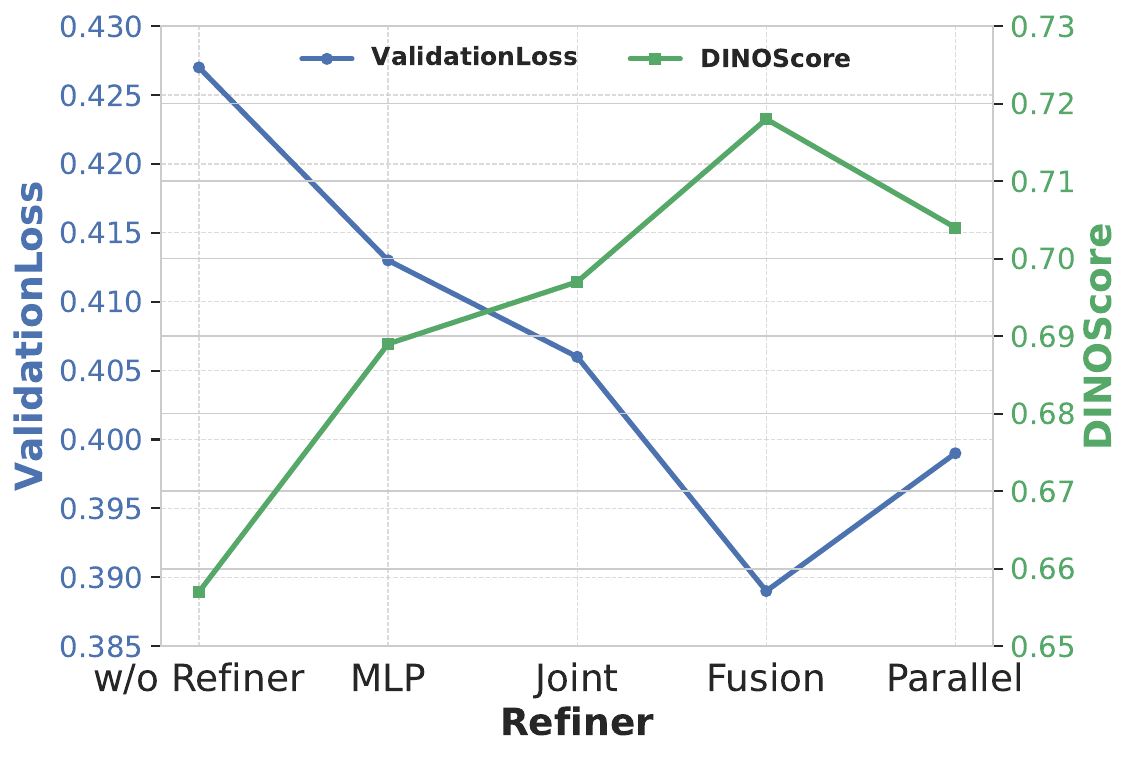}
        \caption{Embedding Refiners}
        \label{fig:refiner_enhancer_choice_a}
    \end{subfigure}
    \hfill
    \begin{subfigure}{0.49\linewidth}
        \centering
        \includegraphics[width=\linewidth]{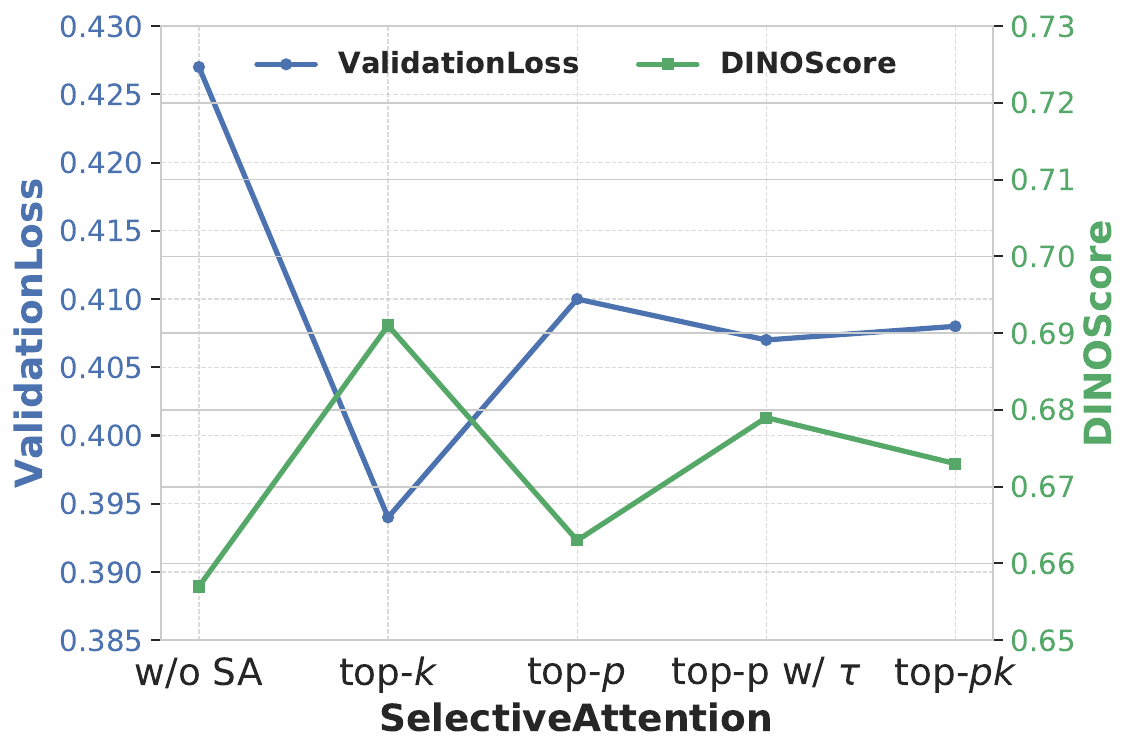}
        \caption{Selective Attention}
        \label{fig:refiner_enhancer_choice_b}
    \end{subfigure}
    \caption{Impact of different architectural choices on validation loss and DINO score. (a) Comparison of various embedding refiners. (b) Comparison of various selective attention.}
    \label{fig:refiner_enhancer_choice}
\end{figure}


\subsection{Unified Multi-modal Representation}
\label{subsec:refiner}
Many diffusion models extract the representation of text and image conditions via separated pre-trained encoders.
However, the misalignment between modalities, defined as the modality-gap, is still an underexplored problem.
Existing work~\cite{liang2022mind,ramasinghe2024accept} that seeks to directly minimize the distance metric between modalities often results in an arbitrary alignment.
We argue that the strategy of minimizing spatial proximity between modalities adversely impacts the nuanced information within visual and linguistic data.
For example, the visual appearance of clothing image cannot be precisely described and aligned with text caption.


To explore the representation of multi-modal conditions, we introduce the Embedding Refiner to align the representation space while keeping the intrinsic property of modality.
Specifically, the Embedding Refiner strives for spatial alignment in the latent space by leveraging learnable feature shifting, and maintains the modality property by masked multi-modal attention.
In this way, we design several structures of Embedding Refiner, as shown in ~\cref{fig:prc_framework} (left), based on the basic operators such as MLP, LayerNorm, and masked Attention.


We begin to introduce MLP with LayerNorm to process concatenated text and image embeddings for feature normalization and shifting.
To model the interaction between modalities, we insert an attention module into the intermediate layer of MLP, which induces the structure of \textit{Joint Transformer}.
However, these two structures have one common drawback that direct concatenation of text and image embedding may lead to degraded representations.
Therefore, we propose the \textit{Fusion Transformer} which learns independent representation before modality interaction and subsequently merges the separated representations into unified embedding by shared attention and MLP layers.

In the attention module, we mask the attention map of text-to-image (key-to-query), named Masked Self Attention, to avoid the adverse impact of text prompt on visual prompt.
In this manner, the visual details of garment in model image are fully derived from visual prompt.
We claim that this divide-and-merge structure with masked attention can effectively redistribute the multi-modal embeddings while keeping pivotal information of modalities.
At last, the \textit{Parallel Transformer} processing text and image prompt separately is proposed for comparison.

~\cref{fig:refiner_enhancer_choice_a} presents the validation loss and DINO score of models armed with different Embedding Refiners.
Compared with baseline, \textit{i.e.}, ``w/o Refiner'', the \textit{MLP} structure achieves considerable improvement on validation loss and DINO score.
This may prove the necessity of Embedding Refiner.
By introducing the attention module, the \textit{Joint Transformer} and \textit{Parallel Transformer} further improve the result, which may be due to the information interaction by attention.
Among all variants, the \textit{Fusion Transformer} achieves the best results.
In this design, the modality embeddings are first preprocessed by separated branches, preserving the modality-specific information.
After information interaction, the shared branch unifies the feature space of text and image conditions, leading to easier integration and generalization of multi-modal embeddings.
Therefore, we adopt the \textit{Fusion Transformer} as our embedding refiner.


\subsection{Attention-Enhanced Correlation}
\label{subsec:enhancer}

\noindent\textbf{Selective Attention.}
The Selective Attention aims to enable each noised token to adaptively attend to the most relevant condition tokens for the preservation of pivotal information.
As illustrated in ~\cref{fig:prc_framework} (right), the Selective Attention is designed as a modified attention taking the noised tokens as queries and the condition tokens as keys and values.
Within the Selective Attention, different selection strategies are explored.
Let $Z$ and $C$ be the latent of noised tokens and embedding of multi-modal condition, respectively.
In Selective Attention, we set the attention query $Q=Z$, and the key/value $K=C$, $V=C$.
Then we calculate the similarity matrix $S = QK^\top$, and apply the softmax with selection strategy $sel\_softmax(\cdot)$ on the similarity matrix.
Thus, the Selective Attention can be implemented as:

\begin{equation}
    selective\_attention(Q,K,V) = sel\_softmax(\frac{QK^\top}{\sqrt{d} \tau})V
\end{equation}
where $d$ and $\tau$ are the feature dimension and pre-defined temperature, respectively.
In our implementation, we explore the selection strategy of top-$k$, top-$p$, and their variants after softmax.
Particularly, we evaluate the impact of temperature by top-$p$ w/ $\tau$, and combination strategy top-$pk$ by selecting the minimal number of tokens with combination of top-$p$ and top-$k$.
By default, we set $\tau=1$.
For the strategies of top-$k$, top-$p$, and top-$pk$, we grid-search the hyper-parameters and select the best value with $k=8$, $p=0.2$, top-$p$ w/ $\tau=0.8$.

~\cref{fig:refiner_enhancer_choice_b} presents the performance of various Selective Attention.
The baseline FLUX ``w/o Selective Attention" yields the highest validation loss and lowest DINO score, indicating the necessity of Selective Attention.
However, the top-$p$ strategy and its variants achieve moderate improvement.
By analyzing the selection tokens, we find that the top-$p$ strategy and its variants usually recall irrelevant condition tokens.
Therefore, we use top-$k$ attention as the Selective Attention in our UMC framework.
~\cref{fig:topk_attn} visualizes the top-$k$ attention mechanism, showing how noised tokens attend to relevant multi-modal tokens from images and texts, thus enhance condition consistency.

\noindent\textbf{Masked Attention.}
The attention mechanism usually calculates the similarity vector along all keys for each query, this may cause information pollution in some situations.
For example, in the consistent generation task, the appearance of subject in reference image should be strictly retained, which may be contaminated by text prompt in attention.
In diffusion model, the condition embeddings can be considered as the anchor information, and thus the attention of noised tokens to condition embeddings should be inhibited.
Therefore, we mask the attention map of text-to-image and noise-to-condition (key-to-query) in Prompt Refiner and FLUX attention, respectively.
These two strategies greatly improve the reliability of condition embeddings in prompt representation and integration stages.


%% file: sec/5_exp.tex
\section{Experiments}
\label{sec:exp}

\begin{figure*}[t]
    \centering
    \includegraphics[width=\linewidth]{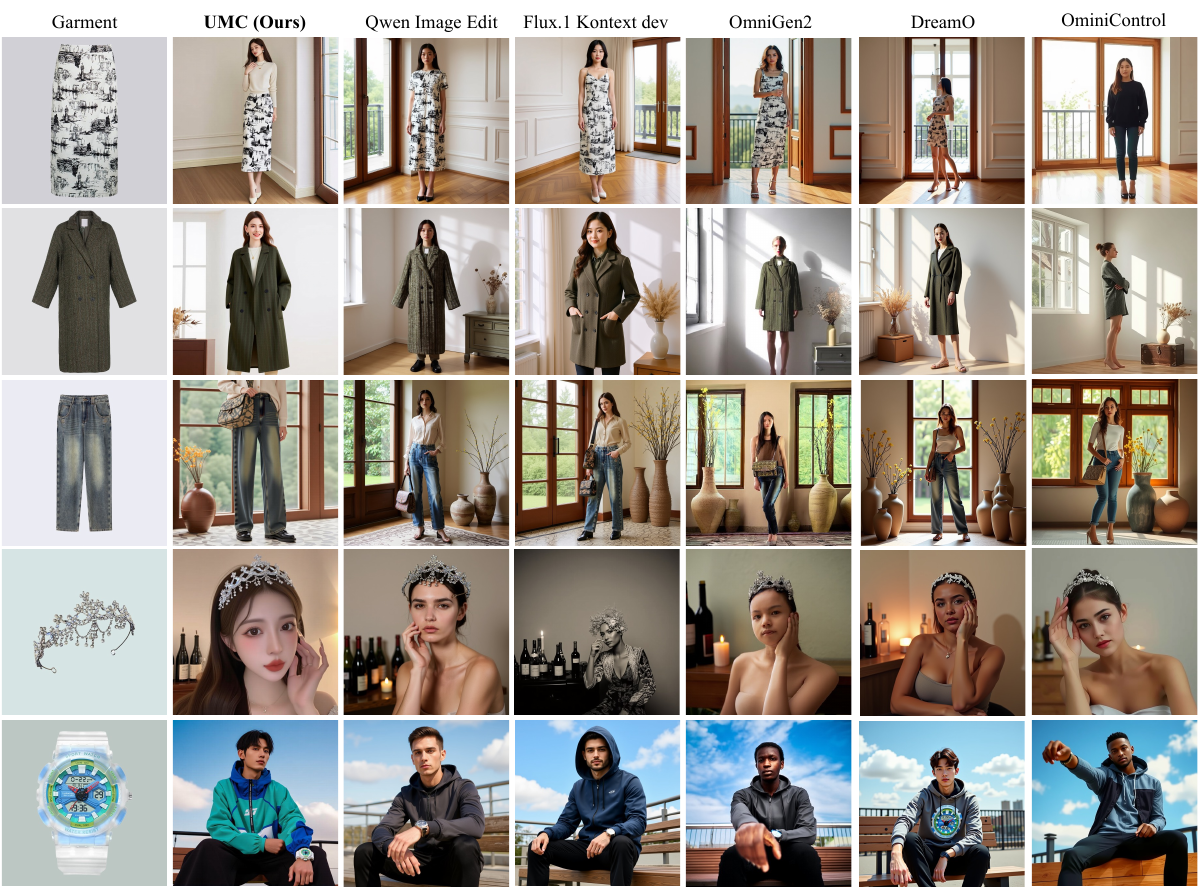}
    \caption{Qualitative results compared with various methods for fashion outfit generation on the test set of Fashion130K}
    \label{fig:compare}
\end{figure*}

\subsection{Experimental Setup}

\noindent\textbf{Implementation Details.}
We use FLUX.1 dev \cite{batifol2025flux} as the pretrained model and fine-tune it via LoRA \cite{hu2022lora} (rank=128). 
Training is conducted in two stages: the first stage updates only the \emph{Embedding Refiner} for 10k steps with all other modules frozen; the second stage jointly trains the \emph{Embedding Refiner}, \emph{Redux}, and \emph{FLUX LoRA} for an additional 30k steps.
We adopt a global batch size of 64 and use Adam optimizer with a learning rate of $10^{-5}$ for both stages. 
To better align with image size distributions of Fashion130K, we employ a multi-resolution bucketing strategy by aspect ratio, grouping samples into three buckets (1:1, 3:4 and 2:3). 
All models are trained on 32 Ascend 910B NPUs using the Fashion130K dataset.

\begin{figure}[H]
    \centering
    \includegraphics[width=\linewidth]{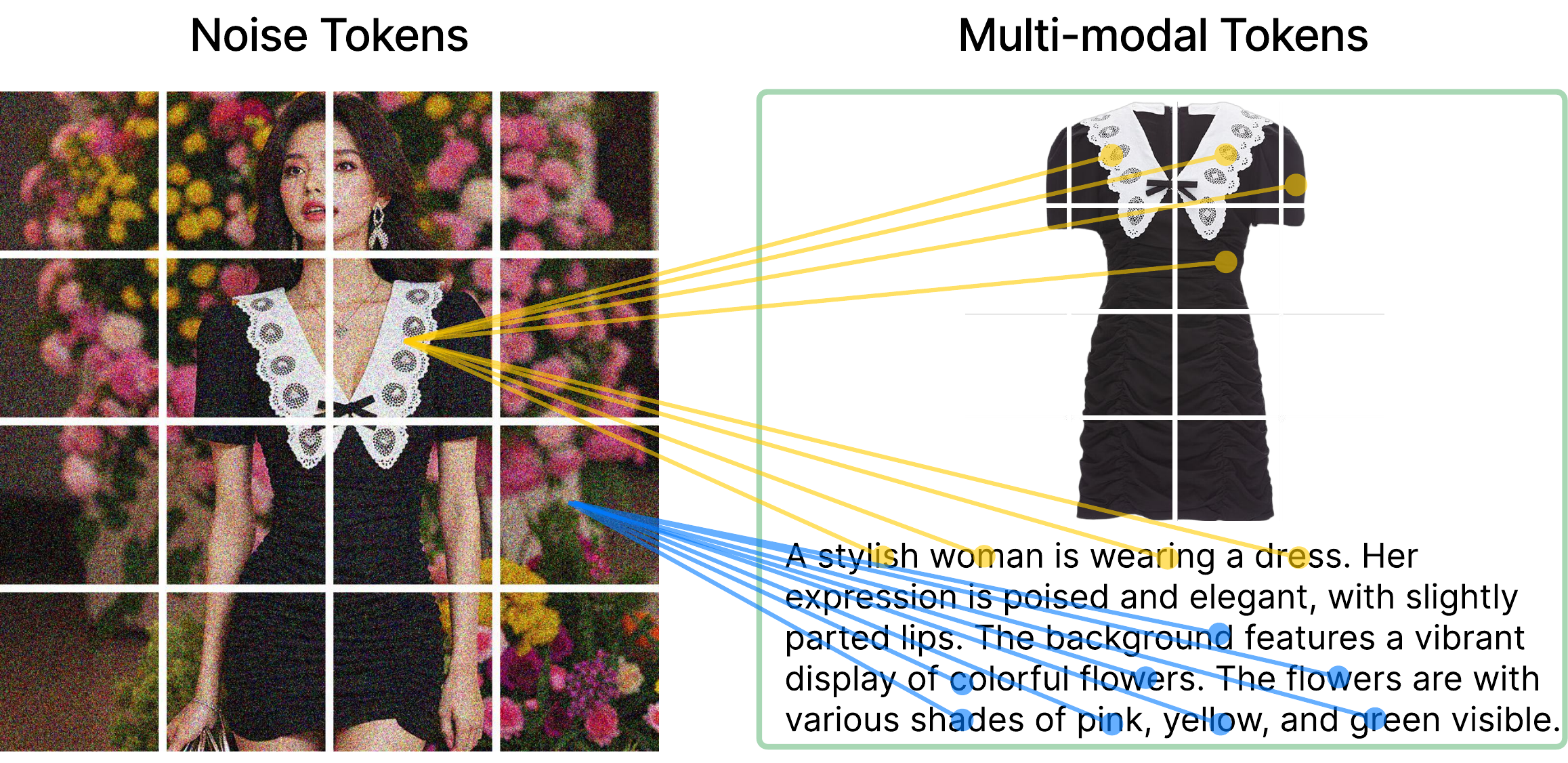}
    \caption{Visualization of Top-$k$ ($k=8$) Attention between noise tokens and multi-modal condition tokens.}
    \label{fig:topk_attn}
\end{figure}

\noindent\textbf{Compared Methods.}
We evaluate our method on the test set of Fashion130K, which contains $3000$ garment-model pairs with rich captions. 
Our evaluation covers three categories of controllable image generation methods: fashion outfit methods (Magic Clothing \cite{chen2024magic}, Any2AnyTryon \cite{guo2025any2anytryon}); subject-driven methods (OmniControl \cite{tan2024ominicontrol}, UNO \cite{wu2025less}, DreamO \cite{mou2025dreamo}, USO \cite{wu2025uso}); and image editing methods (OmniGen \cite{xiao2025omnigen}, ACE++ \cite{mao2025ace++}, OmniGen2 \cite{wu2025omnigen2}, FLUX.1 Kontext Dev \cite{batifol2025flux}, Qwen Image Edit \cite{wu2025qwen}).

\noindent\textbf{Evaluation Metrics.}
We adopt four standard metrics to evaluate visual consistency and text alignment. 
(1) LPIPS \cite{zhang2018unreasonable}: Perceptual distance between a generated image and its reference using deep features with learned channel weights.
(2) DINO: Cosine similarity is computed between generated and reference images using features from DINOv2 \cite{oquab2023dinov2}, capturing visual consistency and identity from different embedding spaces. 
(3) FFA \cite{kotar2023these}: Compare the generated/reference pair via object-intrinsic representations and report the alignment similarity
(4) CLIP-T: Cosine similarity between generated image embeddings and text embeddings from CLIP, measures image-text alignment. 

\begin{table*}[h]
\centering
\setlength{\tabcolsep}{3pt}
\renewcommand{\arraystretch}{1.05}
\begin{tabular}{@{}l c ccccc ccccc@{}}
\toprule
\multirow{2}{*}{\textbf{Method}} & \multirow{2}{*}{\textbf{\# Params*}} & \multicolumn{4}{c}{\textbf{Fashion130K}} & \multicolumn{3}{c}{\textbf{VITON-HD}} \\
\cmidrule(lr){3-6}\cmidrule(lr){7-9}
 & & \textbf{LPIPS ↓} & \textbf{DINO ↑} & \textbf{FFA ↑} & \textbf{CLIP-T ↑} & \textbf{LPIPS ↓} & \textbf{DINO ↑} & \textbf{FFA ↑} \\
\midrule
\rowcolor{gray!12}
\multicolumn{9}{@{}l@{}}{\textit{Subject-Driven Generation Methods}} \\
OminiControl \cite{tan2024ominicontrol}      & 12B & 0.683 & 0.577 & 0.691 & \underline{0.336} & 0.594 & 0.251 & 0.410 \\
UNO \cite{wu2025less}                        & 12B & 0.682 & 0.551 & 0.686 & 0.316 & 0.674 & 0.401 & 0.601 \\
USO \cite{wu2025uso}                         & 12B & 0.656 & 0.538 & 0.647 & \textbf{0.343} & 0.585 & 0.367 & 0.481 \\
DreamO \cite{mou2025dreamo}                  & 12B & 0.657 & 0.628 & 0.765 & 0.332 & 0.711 & 0.396 & 0.697 \\

\midrule
\rowcolor{gray!12}
\multicolumn{9}{@{}l@{}}{\textit{Image Editing Methods}} \\
OmniGen \cite{xiao2025omnigen}               & 3.8B & 0.733 & 0.394 & 0.469 & 0.311 & 0.518 & 0.392 & 0.525 \\
OmniGen2 \cite{wu2025omnigen2}               & 4B & 0.669 & 0.592 & 0.707 & 0.333 & 0.542 & 0.414 & 0.567 \\
ACE++ \cite{mao2025ace++}                    & 12B & 0.666 & 0.604 & 0.740 & 0.329 & 0.604 & 0.376 & 0.561 \\
FLUX.1 Kontext Dev \cite{batifol2025flux}    & 12B & 0.646 & 0.620 & 0.763 & 0.326 & 0.501 & 0.440 & 0.579 \\
Qwen Image Edit \cite{wu2025qwen}            & 20B & \underline{0.634} & \underline{0.673} & \underline{0.805} & 0.326 & 0.500 & \textbf{0.730} & \underline{0.828} \\

\midrule
\rowcolor{gray!12}
\multicolumn{9}{@{}l@{}}{\textit{Outfit Generation Methods}} \\
Magic Clothing \cite{chen2024magic}          & 0.9B & 0.721 & 0.603 & 0.377 & 0.254 & 0.552 & 0.670 & 0.812 \\
Any2AnyTryon \cite{guo2025any2anytryon}      & 12B & 0.658 & 0.551 & 0.664 & 0.331 & \underline{0.481} & 0.633 & 0.755 \\
\textbf{UMC}                         & 12B & \textbf{0.623} & \textbf{0.677} & \textbf{0.818} & 0.332 & 0.515 / \textbf{0.476} & 0.680 / \underline{0.727} & 0.812 / \textbf{0.835} \\

\bottomrule
\end{tabular}
\caption{
Comparison of quantitative results on Fashion130K and VITON-HD.
\textbf{Bold numbers} denote the best performance for each metric within a dataset, while \underline{underlined numbers} indicate the second-best results.
* refers to the number of parameters allocated for image generation.
“/” indicates the results after continued training of UMC on the VITON-HD training set.
}
\label{tab:main_comparison}
\end{table*}

\subsection{Comparison Results}

\noindent\textbf{Qualitative Results.}
As demonstrated in the first row of ~\cref{fig:compare}, only UMC reproduces the correct skirt design. 
Qwen Image Edit \cite{wu2025qwen}, FLUX.1 Kontext Dev \cite{batifol2025flux}, OmniGen2 \cite{wu2025omnigen2}, and DreamO \cite{mou2025dreamo} all collapse to a one-piece dress, indicating weaker control over garment type.
In the third row, UMC renders denim jeans with faithful wash and tone, whereas Qwen Image Edit  \cite{wu2025qwen} shows a color shift, and other methods exhibit shape/style drift.
Specifically, UMC correctly places and shapes the hair accessory (row 4) to match the reference, while competing methods generate mismatched or deformed ornaments.
Overall, UMC demonstrates superior correctness and visual fidelity.

\noindent\textbf{Quantitative Results.}
In ~\cref{tab:main_comparison}, UMC outperforms previous methods across three of four metrics on Fashion130K, achieving the best LPIPS, DINO and FFA score, demonstrating superior preservation of visual details and image-text alignment.
Compared with the strongest subject-driven baselines, UMC reduces LPIPS by 0.033 and improves DINO score by 0.049. 
Against image-editing methods, UMC outperforms the Qwen Image Edit \cite{wu2025qwen} on LPIPS (0.623 vs. 0.634) and FFA (0.818 vs. 0.805), indicating that UMC has better realism and higher object-intrinsic alignment.
Within fashion outfit generation approaches, UMC dominates across all metrics.
These gains indicate that UMC delivers more realistic images (LPIPS), stronger garment consistency (DINO/FFA), and competitive text alignment (CLIP-T).

The performance of UMC on VITON-HD demonstrates the strong generalization capability of the model trained on Fashion130K, while the superior results of the fine-tuned version indicate that the high-quality Fashion130K dataset serves as an effective foundation for post-training. 

\begin{table}[tb]
    \centering
    \begin{tabular}{@{}lccc@{}}
        \toprule
        \textbf{$k$ value} & \textbf{DINO ↑} & \textbf{FFA ↑} & \textbf{CLIP-T ↑} \\
        \midrule
        $k=4$               & 0.641 & 0.774 & 0.320 \\
        $k=8$               & 0.654 & 0.786 & 0.325 \\
        $k=16$              & 0.653 & 0.788 & 0.325 \\
        $k=32$              & 0.649 & 0.788 & 0.326 \\
        $k=4$ w/ ER      & 0.667 & 0.809 & 0.329 \\
        $k=8$ w/ ER      & \textbf{0.674} & \textbf{0.812} & \underline{0.331} \\
        $k=16$ w/ ER     & \textbf{0.674} & \underline{0.810} & \textbf{0.332} \\
        $k=32$ w/ ER     & \underline{0.671} & 0.807 & \underline{0.331} \\
        \bottomrule
    \end{tabular}
    \caption{The impact with and without the Embedding Refiner (ER) of different $k$ values in top-$k$ attention.}
    \label{tab:topk_eval}
\end{table}

\subsection{Ablation Study}

\noindent\textbf{Effect of $k$ in Selective Attention.}
~\cref{tab:topk_eval} shows the impact of different $k$ values ($k=4,8,16,32$). Without Embedding Refiner (ER), $k$=8 achieves optimal performance, suggesting that eight condition tokens per noised token is effective enough to balance the relevance and diversity. Smaller $k$ values (\textit{e.g.}, 4) result in insufficient selection of important condition tokens. In contrast, larger $k$ values (\textit{e.g.}, 16, 32) introduce redundant or less relevant tokens, leading to interference and a reduced ability to maintain precise guidance. Incorporating the ER further elevates performance, again with $k=8$ yielding the best result. Therefore, $k=8$ is selected as a suitable choice for keeping pivotal conditions.

\noindent\textbf{Effect of the Proposed Modules.}
~\cref{tab:eval_module} evaluates the effectiveness of the individual modules and their combinations. The baseline without any proposed modules yields the lowest performance. Using Embedding Refiner (ER) alone notably increases all three metrics, showing its crucial role in fusing visual and text embeddings and optimizing the modality gap. The combination of ER and Selective Attention (SA) further elevates DINO and FFA scores, indicating enhanced subject consistency through their synergy. The Masked Attention (MA) also contributes consistent gains across settings by filtering noised image tokens. Finally, the UMC combining ER, SA, and MA achieves the best results.

\begin{table}[htbp]
    \centering
    \begin{tabular}{@{}lccc@{}}
        \toprule
        \textbf{Method} & \textbf{DINO ↑} & \textbf{FFA ↑} & \textbf{CLIP-T ↑} \\
        \midrule
        baseline        & 0.624 & 0.713 & 0.320 \\
        \midrule
        + MA            & 0.638 & 0.749 & 0.320 \\
        + SA            & 0.654 & 0.786 & 0.321 \\
        + ER            & 0.661 & 0.793 & 0.323 \\
        + SA + MA       & 0.659 & 0.797 & 0.319 \\
        + ER + MA       & 0.668 & 0.802 & 0.323 \\
        + ER + SA       & \underline{0.674} & \underline{0.812} & \underline{0.331} \\
        \midrule
        \textbf{UMC}    & \textbf{0.677} & \textbf{0.818} &\textbf{0.332} \\
        \bottomrule
    \end{tabular}
    \caption{Evaluating the effectiveness of each proposed module on Fashion130K. MA: Masked Attention, SA: Selective Attention, ER: Embedding Refiner.}
    \label{tab:eval_module}
\end{table}


%% file: sec/6_conclusion.tex
\section{Conclusion}
\label{sec:conclusion}

In this paper, we presented UMC, a fashion outfit generation framework that achieves consistent garment transfer under multi-modal condition.
By introducing the Embedding Refiner, we effectively unify visual and text embeddings, where the Fusion Transformer appropriately aligns the representation space without deteriorating the nuanced information of visual prompt. 
Furthermore, the Selective Attention significantly enhances the correlation between multi-modal prompts and noised image for precise detail preservation by selecting the pivotal condition tokens.
To drive research in realistic e-commerce settings, we release Fashion130K, the largest open-source outfit generation dataset to date.
The dataset covers diverse categories, detailed caption and supports multi-resolution training.
Extensive experiments on Fashion130K and public benchmark demonstrate that UMC achieves SoTA performance in garment consistency and fidelity.
We believe that UMC and Fashion130K provide a strong foundation for future advances in outfit generation.


%% file: main.bib
@String(TOG= {ACM Trans. Graph.})

@String(ICLR = {Int. Conf. Learn. Represent.})

@String(AAAI = {AAAI})

@String(TOG   = {ACM TOG})

@String(ICLR  = {ICLR})

@article{ho2020denoising,
  title={Denoising diffusion probabilistic models},
  author={Ho, Jonathan and Jain, Ajay and Abbeel, Pieter},
  journal={Advances in neural information processing systems},
  volume={33},
  pages={6840--6851},
  year={2020}
}

@article{song2020denoising,
  title={Denoising diffusion implicit models},
  author={Song, Jiaming and Meng, Chenlin and Ermon, Stefano},
  journal={arXiv preprint arXiv:2010.02502},
  year={2020}
}

@inproceedings{rombach2022sd,
  title={High-resolution image synthesis with latent diffusion models},
  author={Rombach, Robin and Blattmann, Andreas and Lorenz, Dominik and Esser, Patrick and Ommer, Bj{\"o}rn},
  booktitle={Proceedings of the IEEE/CVF conference on computer vision and pattern recognition},
  pages={10684--10695},
  year={2022}
}

@article{ye2023ip,
  title={Ip-adapter: Text compatible image prompt adapter for text-to-image diffusion models},
  author={Ye, Hu and Zhang, Jun and Liu, Sibo and Han, Xiao and Yang, Wei},
  journal={arXiv preprint arXiv:2308.06721},
  year={2023}
}

@inproceedings{xu2025ootdiffusion,
  title={Ootdiffusion: Outfitting fusion based latent diffusion for controllable virtual try-on},
  author={Xu, Yuhao and Gu, Tao and Chen, Weifeng and Chen, Arlene},
  booktitle={Proceedings of the AAAI Conference on Artificial Intelligence},
  volume={39},
  number={9},
  pages={8996--9004},
  year={2025}
}

@inproceedings{ruiz2023dreambooth,
  title={Dreambooth: Fine tuning text-to-image diffusion models for subject-driven generation},
  author={Ruiz, Nataniel and Li, Yuanzhen and Jampani, Varun and Pritch, Yael and Rubinstein, Michael and Aberman, Kfir},
  booktitle={Proceedings of the IEEE/CVF conference on computer vision and pattern recognition},
  pages={22500--22510},
  year={2023}
}

@inproceedings{xiao2025omnigen,
  title={Omnigen: Unified image generation},
  author={Xiao, Shitao and Wang, Yueze and Zhou, Junjie and Yuan, Huaying and Xing, Xingrun and Yan, Ruiran and Li, Chaofan and Wang, Shuting and Huang, Tiejun and Liu, Zheng},
  booktitle={Proceedings of the Computer Vision and Pattern Recognition Conference},
  pages={13294--13304},
  year={2025}
}

@article{wu2025less,
  title={Less-to-more generalization: Unlocking more controllability by in-context generation},
  author={Wu, Shaojin and Huang, Mengqi and Wu, Wenxu and Cheng, Yufeng and Ding, Fei and He, Qian},
  journal={arXiv preprint arXiv:2504.02160},
  year={2025}
}

@article{wu2025uso,
  title={Uso: Unified style and subject-driven generation via disentangled and reward learning},
  author={Wu, Shaojin and Huang, Mengqi and Cheng, Yufeng and Wu, Wenxu and Tian, Jiahe and Luo, Yiming and Ding, Fei and He, Qian},
  journal={arXiv preprint arXiv:2508.18966},
  year={2025}
}

@inproceedings{peebles2023dit,
  title={Scalable diffusion models with transformers},
  author={Peebles, William and Xie, Saining},
  booktitle={Proceedings of the IEEE/CVF international conference on computer vision},
  pages={4195--4205},
  year={2023}
}

@article{podell2023sdxl,
  title={Sdxl: Improving latent diffusion models for high-resolution image synthesis},
  author={Podell, Dustin and English, Zion and Lacey, Kyle and Blattmann, Andreas and Dockhorn, Tim and M{\"u}ller, Jonas and Penna, Joe and Rombach, Robin},
  journal={arXiv preprint arXiv:2307.01952},
  year={2023}
}

@article{esser2024sd3,
  title={Scaling Rectified Flow Transformers for High-Resolution Image Synthesis},
  author={Esser, Patrick and Kulal, Sumith and Blattmann, Andreas and Entezari, Rahim and M{\"u}ller, Jonas and Saini, Harry and Levi, Yam and Lorenz, Dominik and Sauer, Axel and Boesel, Frederic and others},
  journal={arXiv preprint arXiv:2403.03206},
  year={2024}
}

@article{batifol2025flux,
  title={FLUX. 1 Kontext: Flow Matching for In-Context Image Generation and Editing in Latent Space},
  author={Batifol, Stephen and Blattmann, Andreas and Boesel, Frederic and Consul, Saksham and Diagne, Cyril and Dockhorn, Tim and English, Jack and English, Zion and Esser, Patrick and Kulal, Sumith and others},
  journal={arXiv e-prints},
  pages={arXiv--2506},
  year={2025}
}

@article{wu2025qwen,
  title={Qwen-image technical report},
  author={Wu, Chenfei and Li, Jiahao and Zhou, Jingren and Lin, Junyang and Gao, Kaiyuan and Yan, Kun and Yin, Sheng-ming and Bai, Shuai and Xu, Xiao and Chen, Yilei and others},
  journal={arXiv preprint arXiv:2508.02324},
  year={2025}
}

@article{mou2025dreamo,
  title={DreamO: A Unified Framework for Image Customization},
  author={Mou, Chong and Wu, Yanze and Wu, Wenxu and Guo, Zinan and Zhang, Pengze and Cheng, Yufeng and Luo, Yiming and Ding, Fei and Zhang, Shiwen and Li, Xinghui and others},
  journal={arXiv preprint arXiv:2504.16915},
  year={2025}
}

@inproceedings{chen2024magic,
  title={Magic clothing: Controllable garment-driven image synthesis},
  author={Chen, Weifeng and Gu, Tao and Xu, Yuhao and Chen, Arlene},
  booktitle={Proceedings of the 32nd ACM International Conference on Multimedia},
  pages={6939--6948},
  year={2024}
}

@inproceedings{guo2025any2anytryon,
  title={Any2anytryon: Leveraging adaptive position embeddings for versatile virtual clothing tasks},
  author={Guo, Hailong and Zeng, Bohan and Song, Yiren and Zhang, Wentao and Liu, Jiaming and Zhang, Chuang},
  booktitle={Proceedings of the IEEE/CVF International Conference on Computer Vision},
  pages={19085--19096},
  year={2025}
}

@article{goodfellow2014generative,
  title={Generative adversarial nets},
  author={Goodfellow, Ian J and Pouget-Abadie, Jean and Mirza, Mehdi and Xu, Bing and Warde-Farley, David and Ozair, Sherjil and Courville, Aaron and Bengio, Yoshua},
  journal={Advances in neural information processing systems},
  volume={27},
  year={2014}
}

@article{ramesh2022dalle2,
  title={Hierarchical text-conditional image generation with clip latents},
  author={Ramesh, Aditya and Dhariwal, Prafulla and Nichol, Alex and Chu, Casey and Chen, Mark},
  journal={arXiv preprint arXiv:2204.06125},
  volume={1},
  number={2},
  pages={3},
  year={2022}
}

@article{chen2023pixart,
  title={Pixart-alpha: Fast training of diffusion transformer for photorealistic text-to-image synthesis},
  author={Chen, Junsong and Yu, Jincheng and Ge, Chongjian and Yao, Lewei and Xie, Enze and Wu, Yue and Wang, Zhongdao and Kwok, James and Luo, Ping and Lu, Huchuan and others},
  journal={arXiv preprint arXiv:2310.00426},
  year={2023}
}

@article{saharia2022photorealistic,
  title={Photorealistic text-to-image diffusion models with deep language understanding},
  author={Saharia, Chitwan and Chan, William and Saxena, Saurabh and Li, Lala and Whang, Jay and Denton, Emily L and Ghasemipour, Kamyar and Gontijo Lopes, Raphael and Karagol Ayan, Burcu and Salimans, Tim and others},
  journal={Advances in neural information processing systems},
  volume={35},
  pages={36479--36494},
  year={2022}
}

@article{baldridge2024imagen,
  title={Imagen 3},
  author={Baldridge, Jason and Bauer, Jakob and Bhutani, Mukul and Brichtova, Nicole and Bunner, Andrew and Castrejon, Lluis and Chan, Kelvin and Chen, Yichang and Dieleman, Sander and Du, Yuqing and others},
  journal={arXiv preprint arXiv:2408.07009},
  year={2024}
}

@article{qin2025lumina,
  title={Lumina-image 2.0: A unified and efficient image generative framework},
  author={Qin, Qi and Zhuo, Le and Xin, Yi and Du, Ruoyi and Li, Zhen and Fu, Bin and Lu, Yiting and Yuan, Jiakang and Li, Xinyue and Liu, Dongyang and others},
  journal={arXiv preprint arXiv:2503.21758},
  year={2025}
}

@article{team2024kolors,
  title={Kolors: Effective training of diffusion model for photorealistic text-to-image synthesis},
  author={K, Team},
  journal={arXiv preprint},
  year={2024}
}

@article{li2024hunyuan,
  title={Hunyuan-dit: A powerful multi-resolution diffusion transformer with fine-grained chinese understanding},
  author={Li, Zhimin and Zhang, Jianwei and Lin, Qin and Xiong, Jiangfeng and Long, Yanxin and Deng, Xinchi and Zhang, Yingfang and Liu, Xingchao and Huang, Minbin and Xiao, Zedong and others},
  journal={arXiv preprint arXiv:2405.08748},
  year={2024}
}

@article{guo2024pulid,
  title={Pulid: Pure and lightning id customization via contrastive alignment},
  author={Guo, Zinan and Wu, Yanze and Zhuowei, Chen and Zhang, Peng and He, Qian and others},
  journal={Advances in neural information processing systems},
  volume={37},
  pages={36777--36804},
  year={2024}
}

@inproceedings{zhou2025learning,
  title={Learning flow fields in attention for controllable person image generation},
  author={Zhou, Zijian and Liu, Shikun and Han, Xiao and Liu, Haozhe and Ng, Kam Woh and Xie, Tian and Cong, Yuren and Li, Hang and Xu, Mengmeng and P{\'e}rez-R{\'u}a, Juan-Manuel and others},
  booktitle={Proceedings of the Computer Vision and Pattern Recognition Conference},
  pages={2491--2501},
  year={2025}
}

@article{choi2024improving,
  title={Improving diffusion models for virtual try-on},
  author={Choi, Yisol and Kwak, Sangkyung and Lee, Kyungmin and Choi, Hyungwon and Shin, Jinwoo},
  journal={arXiv e-prints},
  pages={arXiv--2403},
  year={2024}
}

@article{wang2024stablegarment,
  title={Stablegarment: Garment-centric generation via stable diffusion},
  author={Wang, Rui and Guo, Hailong and Liu, Jiaming and Li, Huaxia and Zhao, Haibo and Tang, Xu and Hu, Yao and Tang, Hao and Li, Peipei},
  journal={arXiv preprint arXiv:2403.10783},
  year={2024}
}

@article{li2024anyfit,
  title={Anyfit: Controllable virtual try-on for any combination of attire across any scenario},
  author={Li, Yuhan and Zhou, Hao and Shang, Wenxiang and Lin, Ran and Chen, Xuanhong and Ni, Bingbing},
  journal={arXiv preprint arXiv:2405.18172},
  year={2024}
}

@article{lipman2022flow,
  title={Flow matching for generative modeling},
  author={Lipman, Yaron and Chen, Ricky TQ and Ben-Hamu, Heli and Nickel, Maximilian and Le, Matt},
  journal={arXiv preprint arXiv:2210.02747},
  year={2022}
}

@article{liu2022flow,
  title={Flow straight and fast: Learning to generate and transfer data with rectified flow},
  author={Liu, Xingchao and Gong, Chengyue and Liu, Qiang},
  journal={arXiv preprint arXiv:2209.03003},
  year={2022}
}

@article{liang2022mind,
  title={Mind the gap: Understanding the modality gap in multi-modal contrastive representation learning},
  author={Liang, Victor Weixin and Zhang, Yuhui and Kwon, Yongchan and Yeung, Serena and Zou, James Y},
  journal={Advances in Neural Information Processing Systems},
  volume={35},
  pages={17612--17625},
  year={2022}
}

@article{hu2022lora,
  title={Lora: Low-rank adaptation of large language models.},
  author={Hu, Edward J and Shen, Yelong and Wallis, Phillip and Allen-Zhu, Zeyuan and Li, Yuanzhi and Wang, Shean and Wang, Lu and Chen, Weizhu and others},
  journal={ICLR},
  volume={1},
  number={2},
  pages={3},
  year={2022}
}

@article{tan2024ominicontrol,
  title={Ominicontrol: Minimal and universal control for diffusion transformer},
  author={Tan, Zhenxiong and Liu, Songhua and Yang, Xingyi and Xue, Qiaochu and Wang, Xinchao},
  journal={arXiv preprint arXiv:2411.15098},
  year={2024}
}

@article{mao2025ace++,
  title={Ace++: Instruction-based image creation and editing via context-aware content filling},
  author={Mao, Chaojie and Zhang, Jingfeng and Pan, Yulin and Jiang, Zeyinzi and Han, Zhen and Liu, Yu and Zhou, Jingren},
  journal={arXiv preprint arXiv:2501.02487},
  year={2025}
}

@article{wu2025omnigen2,
  title={OmniGen2: Exploration to Advanced Multimodal Generation},
  author={Wu, Chenyuan and Zheng, Pengfei and Yan, Ruiran and Xiao, Shitao and Luo, Xin and Wang, Yueze and Li, Wanli and Jiang, Xiyan and Liu, Yexin and Zhou, Junjie and others},
  journal={arXiv preprint arXiv:2506.18871},
  year={2025}
}

@article{oquab2023dinov2,
  title={Dinov2: Learning robust visual features without supervision},
  author={Oquab, Maxime and Darcet, Timoth{\'e}e and Moutakanni, Th{\'e}o and Vo, Huy and Szafraniec, Marc and Khalidov, Vasil and Fernandez, Pierre and Haziza, Daniel and Massa, Francisco and El-Nouby, Alaaeldin and others},
  journal={arXiv preprint arXiv:2304.07193},
  year={2023}
}

@inproceedings{karras2019style,
  title={A style-based generator architecture for generative adversarial networks},
  author={Karras, Tero and Laine, Samuli and Aila, Timo},
  booktitle={Proceedings of the IEEE/CVF conference on computer vision and pattern recognition},
  pages={4401--4410},
  year={2019}
}

@article{brock2018large,
  title={Large scale GAN training for high fidelity natural image synthesis},
  author={Brock, Andrew and Donahue, Jeff and Simonyan, Karen},
  journal={arXiv preprint arXiv:1809.11096},
  year={2018}
}

@inproceedings{li2025dual,
  title={Dual diffusion for unified image generation and understanding},
  author={Li, Zijie and Li, Henry and Shi, Yichun and Farimani, Amir Barati and Kluger, Yuval and Yang, Linjie and Wang, Peng},
  booktitle={Proceedings of the Computer Vision and Pattern Recognition Conference},
  pages={2779--2790},
  year={2025}
}

@article{holderrieth2025introduction,
  title={An Introduction to Flow Matching and Diffusion Models},
  author={Holderrieth, Peter and Erives, Ezra},
  journal={arXiv preprint arXiv:2506.02070},
  year={2025}
}

@inproceedings{neuberger2020image,
  title={Image based virtual try-on network from unpaired data},
  author={Neuberger, Assaf and Borenstein, Eran and Hilleli, Bar and Oks, Eduard and Alpert, Sharon},
  booktitle={Proceedings of the IEEE/CVF conference on computer vision and pattern recognition},
  pages={5184--5193},
  year={2020}
}

@article{lewis2021tryongan,
  title={Tryongan: Body-aware try-on via layered interpolation},
  author={Lewis, Kathleen M and Varadharajan, Srivatsan and Kemelmacher-Shlizerman, Ira},
  journal={ACM Transactions on Graphics (TOG)},
  volume={40},
  number={4},
  pages={1--10},
  year={2021},
  publisher={ACM New York, NY, USA}
}

@inproceedings{li2021toward,
  title={Toward accurate and realistic outfits visualization with attention to details},
  author={Li, Kedan and Chong, Min Jin and Zhang, Jeffrey and Liu, Jingen},
  booktitle={Proceedings of the IEEE/CVF conference on computer vision and pattern recognition},
  pages={15546--15555},
  year={2021}
}

@inproceedings{yildirim2019generating,
  title={Generating high-resolution fashion model images wearing custom outfits},
  author={Yildirim, Gokhan and Jetchev, Nikolay and Vollgraf, Roland and Bergmann, Urs},
  booktitle={Proceedings of the IEEE/CVF international conference on computer vision workshops},
  pages={0--0},
  year={2019}
}

@inproceedings{choi2021viton,
  title={Viton-hd: High-resolution virtual try-on via misalignment-aware normalization},
  author={Choi, Seunghwan and Park, Sunghyun and Lee, Minsoo and Choo, Jaegul},
  booktitle={Proceedings of the IEEE/CVF conference on computer vision and pattern recognition},
  pages={14131--14140},
  year={2021}
}

@inproceedings{hsieh2019fashionon,
  title={FashionOn: Semantic-guided image-based virtual try-on with detailed human and clothing information},
  author={Hsieh, Chia-Wei and Chen, Chieh-Yun and Chou, Chien-Lung and Shuai, Hong-Han and Liu, Jiaying and Cheng, Wen-Huang},
  booktitle={Proceedings of the 27th ACM international conference on multimedia},
  pages={275--283},
  year={2019}
}

@inproceedings{liu2016deepfashion,
  title={Deepfashion: Powering robust clothes recognition and retrieval with rich annotations},
  author={Liu, Ziwei and Luo, Ping and Qiu, Shi and Wang, Xiaogang and Tang, Xiaoou},
  booktitle={Proceedings of the IEEE conference on computer vision and pattern recognition},
  pages={1096--1104},
  year={2016}
}

@inproceedings{dong2019towards,
  title={Towards multi-pose guided virtual try-on network},
  author={Dong, Haoye and Liang, Xiaodan and Shen, Xiaohui and Wang, Bochao and Lai, Hanjiang and Zhu, Jia and Hu, Zhiting and Yin, Jian},
  booktitle={Proceedings of the IEEE/CVF international conference on computer vision},
  pages={9026--9035},
  year={2019}
}

@inproceedings{zheng2019virtually,
  title={Virtually trying on new clothing with arbitrary poses},
  author={Zheng, Na and Song, Xuemeng and Chen, Zhaozheng and Hu, Linmei and Cao, Da and Nie, Liqiang},
  booktitle={Proceedings of the 27th ACM international conference on multimedia},
  pages={266--274},
  year={2019}
}

@inproceedings{yoo2016pixel,
  title={Pixel-level domain transfer},
  author={Yoo, Donggeun and Kim, Namil and Park, Sunggyun and Paek, Anthony S and Kweon, In So},
  booktitle={European conference on computer vision},
  pages={517--532},
  year={2016},
  organization={Springer}
}

@inproceedings{morelli2022dress,
  title={Dress code: High-resolution multi-category virtual try-on},
  author={Morelli, Davide and Fincato, Matteo and Cornia, Marcella and Landi, Federico and Cesari, Fabio and Cucchiara, Rita},
  booktitle={Proceedings of the IEEE/CVF conference on computer vision and pattern recognition},
  pages={2231--2235},
  year={2022}
}

@inproceedings{wang2025mv,
  title={Mv-vton: Multi-view virtual try-on with diffusion models},
  author={Wang, Haoyu and Zhang, Zhilu and Di, Donglin and Zhang, Shiliang and Zuo, Wangmeng},
  booktitle={Proceedings of the AAAI Conference on Artificial Intelligence},
  volume={39},
  number={7},
  pages={7682--7690},
  year={2025}
}

@inproceedings{han2018viton,
  title={Viton: An image-based virtual try-on network},
  author={Han, Xintong and Wu, Zuxuan and Wu, Zhe and Yu, Ruichi and Davis, Larry S},
  booktitle={Proceedings of the IEEE conference on computer vision and pattern recognition},
  pages={7543--7552},
  year={2018}
}

@article{kotar2023these,
  title={Are these the same apple? comparing images based on object intrinsics},
  author={Kotar, Klemen and Tian, Stephen and Yu, Hong-Xing and Yamins, Dan and Wu, Jiajun},
  journal={Advances in Neural Information Processing Systems},
  volume={36},
  pages={40853--40871},
  year={2023}
}

@inproceedings{zhang2018unreasonable,
  title={The unreasonable effectiveness of deep features as a perceptual metric},
  author={Zhang, Richard and Isola, Phillip and Efros, Alexei A and Shechtman, Eli and Wang, Oliver},
  booktitle={Proceedings of the IEEE conference on computer vision and pattern recognition},
  pages={586--595},
  year={2018}
}

@inproceedings{ramasinghe2024accept,
  title={Accept the modality gap: An exploration in the hyperbolic space},
  author={Ramasinghe, Sameera and Shevchenko, Violetta and Avraham, Gil and Thalaiyasingam, Ajanthan},
  booktitle={Proceedings of the IEEE/CVF Conference on Computer Vision and Pattern Recognition},
  pages={27263--27272},
  year={2024}
}

@article{han2024emma,
  title={Emma: Your text-to-image diffusion model can secretly accept multi-modal prompts},
  author={Han, Yucheng and Wang, Rui and Zhang, Chi and Hu, Juntao and Cheng, Pei and Fu, Bin and Zhang, Hanwang},
  journal={arXiv preprint arXiv:2406.09162},
  year={2024}
}
